\documentclass[letterpaper, 10 pt, conference]{ieeeconf}  %

\IEEEoverridecommandlockouts                              %

\usepackage{amssymb}  %

\usepackage[dvipsnames, table]{xcolor}
\usepackage{amsmath}
\usepackage{breqn}
\usepackage{float}
\usepackage{subcaption, graphicx}
\definecolor{red}{RGB}{255, 0, 0}   %
\definecolor{blue}{RGB}{0, 0, 255}   %
\definecolor{orange}{RGB}{255, 77, 0}   %
\definecolor{green}{RGB}{0, 128, 0}   %
\definecolor{purple}{RGB}{160, 32, 240}   %
\definecolor{lightblue}{RGB}{52, 155, 235}   %
\definecolor{darkmagenta}{RGB}{204, 51, 139}

\usepackage{ifthen}
\newboolean{highlight_additions}
\setboolean{highlight_additions}{false} %
\newcommand{\rev}[1]{%
  \ifthenelse{\boolean{highlight_additions}}%
    {\textcolor{blue}{#1}}%
    {#1}%
}

\usepackage{todonotes}
\usepackage{outlines}
\usepackage[font=footnotesize]{caption}
\usepackage{titlesec}
\AtBeginDocument{
  \setlength\abovedisplayskip{0pt}
  \setlength\belowdisplayskip{0pt}
}

\usepackage{multirow}
\usepackage{hyperref}
\usepackage{cleveref}

\usepackage{booktabs}
\usepackage{xspace}
\usepackage{array}
\usepackage{tabularx}
\usepackage{pifont}%
\usepackage{makecell}

\usepackage{fontawesome}
\usepackage{scalerel}
\usepackage{rotating}
\usepackage{gensymb}

\usepackage[space]{cite}

\usepackage{pgf}
\definecolor{high}{HTML}{228B22}
\definecolor{low}{HTML}{FFFFFF}

\newcommand{\ourName}{DexForce}

\usepackage[scaled=.8]{FiraMono}

\title{\LARGE \bf
\ourName{}: Extracting Force-informed Actions from Kinesthetic Demonstrations for Dexterous Manipulation
}

\author{
Claire Chen$^{1}$, Zhongchun Yu$^{1}$, Hojung Choi$^{1}$, Mark Cutkosky$^{1}$, and Jeannette Bohg$^1$
\thanks{ $^1$Stanford University, CA, USA.}
\thanks{\,\,\,Toyota Research Institute provided funds to support this work.}
}
\begin{document}

\maketitle
\thispagestyle{empty}
\pagestyle{empty}

\begin{abstract}
Imitation learning requires high-quality demonstrations consisting of sequences of state-action pairs. For contact-rich dexterous manipulation tasks that require dexterity, the actions in these state-action pairs must produce the right forces. Current widely-used methods for collecting dexterous manipulation demonstrations are difficult to use for demonstrating contact-rich tasks due to unintuitive human-to-robot motion retargeting and the lack of direct haptic feedback. Motivated by these concerns, we propose \ourName{}. \mbox{\ourName{}} leverages contact forces, measured during kinesthetic demonstrations, to compute force-informed actions for policy learning. 
We collect demonstrations for six tasks and show that policies trained on our force-informed actions achieve an average success rate of 76\% across all tasks. In contrast, policies trained directly on actions that do not account for contact forces have near-zero success rates. We also conduct a study ablating the inclusion of force data in policy observations. We find that while using force data never hurts policy performance, it helps most for tasks that require advanced levels of precision and coordination, like opening an AirPods case and unscrewing a nut. Videos at
\href{https://clairelc.github.io/dexforce.github.io/}{https://clairelc.github.io/dexforce.github.io/}

\end{abstract}

\section{Introduction} \label{sec:intro}

\begin{figure}[t]
    \centering
    \includegraphics[width=\linewidth]{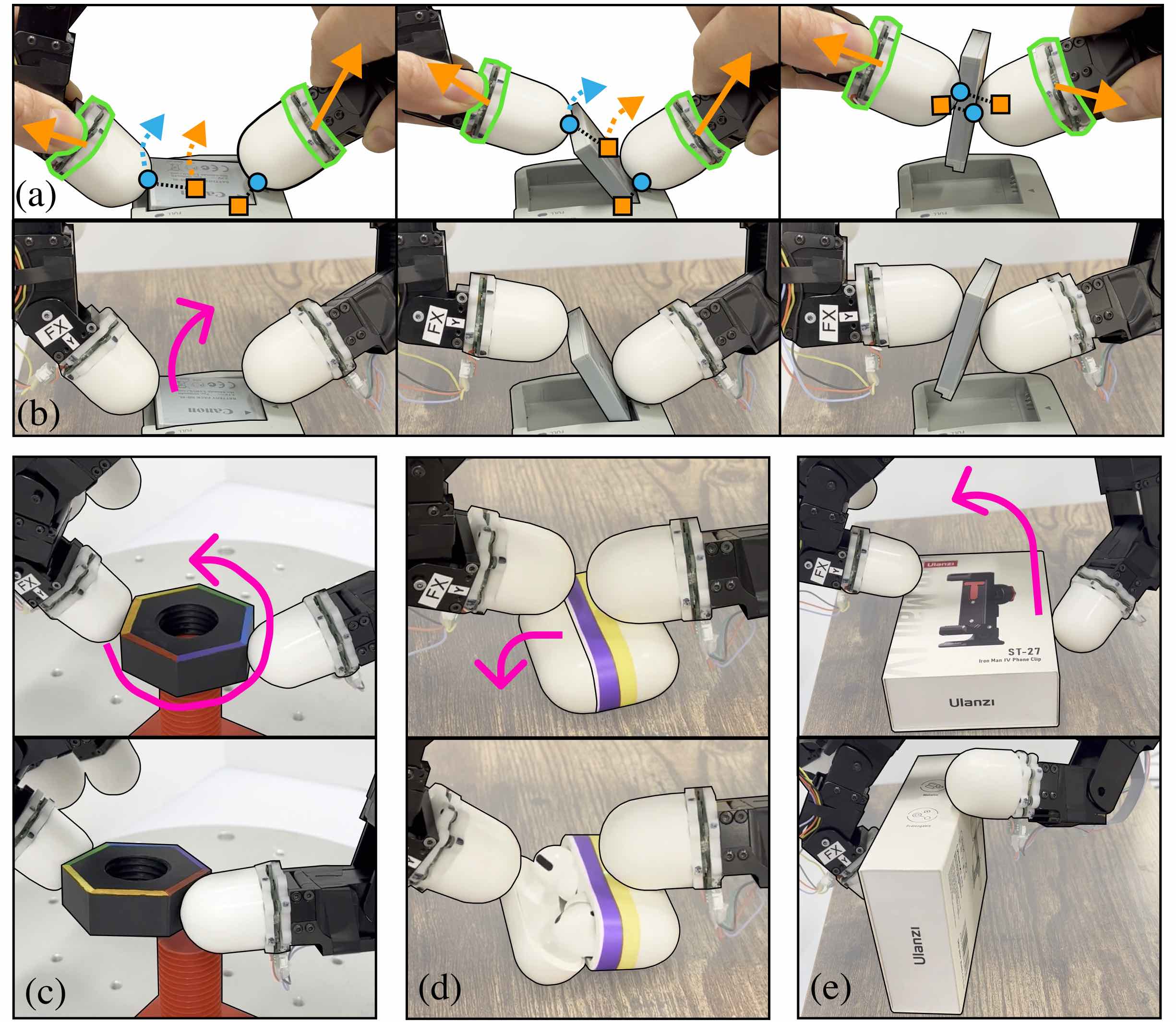}
    \caption{(a) \ourName{} extracts force-informed actions (orange squares) from kinesthetic demonstrations by augmenting observed robot positions (blue circles) according to contact forces (orange arrows) measured with 6-axis force-torque sensors (green). (b) Executing force-informed actions allows the robot to reproduce forces applied in the kinesthetic demonstration, thereby executing the task. \ourName{} enables us to collect high-quality demonstrations for training policies on a variety of contact-rich tasks like (c) unscrewing a nut, (d) opening an AirPods case and (e) flipping a box.}
    \vspace{-6mm}
    \label{fig:teaser}
\end{figure}

Successful dexterous manipulation hinges on applying the right forces, at the right times, in the right contact locations. This is especially important for fine-grained, contact-rich tasks, which demand a more nuanced application of forces than pick-and-place tasks. Consider opening an AirPods case (Fig. \ref{fig:teaser}d) and flipping a box (Fig. \ref{fig:teaser}e), which require applying precise forces at multiple contact points. Other tasks, such as grasping a thin camera battery (Fig. \ref{fig:teaser}b) and unscrewing a nut (Fig. \ref{fig:teaser}c) not only involve applying precise forces, but also require making and breaking contact with the object in a coordinated fashion. To succeed at such tasks, a robot must know how to apply the proper forces.

One way of teaching a robot how to apply the proper forces is through imitation learning \cite{zhao2023aloha, zhao2024alohaunleashedsimplerecipe, chi2023diffusionpolicy}. Critically, successful imitation learning requires high-quality demonstrations consisting of sequences of state-action pairs. \cite{ravichandar2020ilsurvey}.
For contact-rich dexterous manipulation, the actions in these state-action pairs must impart the proper forces.

Dexterous manipulation demonstrations are most commonly collected via teleoperation, where robot actions are obtained by tracking human hand motion and retargeting this to robot positions
\cite{arunachalam2022holodex,cheng2024opentv,handa2019dexpilot, iyer2024openteach, qin2023anyteleop, romero2024eyesighthand, shaw2024bimanual, yang2024ace,ding2024bunnyvisionpro, mannam2024designinganthropomorphicsofthands}. Other works use retargeting to learn directly from human demonstrations \cite{wang2024dexcap, chen2024arcap}.
However, it remains challenging to use retargeted data to obtain demonstrations for tasks that require dexterity. This is due to two inherent limitations. First, a human hand does not move like most robot hands; this kinematic difference, sometimes referred to as the ``human-to-robot correspondence problem'' \cite{si2024tilde, arduengo2021humantorobot}, makes it unintuitive for a human demonstrator to execute precise motions on a robot hand by moving their own hand \cite{wu2023gello}. 
Second, the demonstrator cannot feel what the robot feels during task execution; this lack of haptic feedback makes it especially difficult to execute actions that apply the correct forces to an object \cite{morris2007hapticfeedback, strom2006earlyexposure}.

Motivated by the shortcomings of current dexterous manipulation data collection methods, particularly for contact-rich tasks, we propose \ourName{}. The key element of \ourName{} is how actions are obtained. Given that successful contact-rich manipulation requires actions that apply the right forces, \ourName{} leverages contact forces, measured during kinesthetic demonstrations, to inform actions for policy learning. Kinesthetic teaching, where the demonstrator manually moves a robot to complete a task (Fig. \ref{fig:teaser}a), provides a natural way of demonstrating contact-rich tasks. It not only allows the operator to utilize the full dexterity of the robot hand, but also gives the operator direct haptic feedback.

To record contact forces, we instrument the robot hand with 6-axis force-torque sensors mounted at the base of each fingertip (green in Fig. \ref{fig:teaser}a). However, even when using a robot hand equipped with force sensing, kinesthetic demonstrations only provide state, namely the observed robot motion and measured forces; they do not provide the actions that would enable the robot to reproduce the measured forces.
\ourName{} computes these actions, which we term \textit{force-informed actions}, by augmenting observed robot motion according to the measured forces. When tracked with an impedance controller, these actions allow a robot to generate the right motion and forces to complete the task.

\ourName{} enables us to collect high-quality demonstrations for six contact-rich tasks, which we then use for imitation learning. Policies trained on our force-informed actions achieve an average success rate of 76\% across all six tasks. In contrast, policies trained directly on actions that do not account for contact forces have near-zero success rates. This stark difference in performance underscores that our method of extracting force-informed actions is crucial for learning capable policies. We also study how including force data in policy observations impacts the performance of contact-rich manipulation policies. Our findings show that while including force in the observation improves policy performance for all six tasks, it matters most for tasks that require precision and coordination, like opening an AirPods case and unscrewing a nut.
In summary, we contribute:

1) \ourName{}, a method for collecting demonstrations of contact-rich dexterous manipulation tasks by using contact forces to extract force-informed actions from kinesthetic demonstrations, and

2) a study showing that including forces in policy observations results in larger policy performance improvements for tasks that demand greater precision and coordination.

\section{Related work} \label{sec:related-work}

\noindent \textbf{Collecting dexterous hand demonstrations:}
This work proposes a solution for collecting demonstrations of fine-grained dexterous manipulation, which, as discussed in Section \ref{sec:intro}, has been a limitation of widely-used retargeting-based data collection methods \cite{arunachalam2022holodex,cheng2024opentv,handa2019dexpilot, iyer2024openteach, qin2023anyteleop, romero2024eyesighthand, shaw2024bimanual, yang2024ace,ding2024bunnyvisionpro, mannam2024designinganthropomorphicsofthands,wang2024dexcap, chen2024arcap}. In particular, these methods lack haptic feedback and suffer from the human-to-robot correspondence problem.

Some works propose methods that mitigate the human-to-robot correspondence problem but still lack haptic feedback. ResPilot \cite{naughton2024respilot} improves retargeting-based teleoperation with residual Gaussian Process learning. However, it is designed for finger-gaiting motions, making it difficult to apply to more general dexterity.
Tilde \cite{si2024tilde} uses a twin robot to puppeteer a custom four-fingered hand. Puppeteering is a good way of circumventing the human-to-robot correspondence problem but has a high hardware overhead. Critically, neither work provides haptic feedback. The Tilde authors note that ``tactile feedback could lead to more intentional and controlled interactions with objects in highly-dexterous tasks'' \cite{si2024tilde}. \ourName{} leverages kinesthetic teaching, which not only avoids the human-to-robot correspondence problem, but also provides the user with direct haptic feedback.

In a similar family to kinesthetic teaching, ``hand-over-hand'' or ``exoskeleton'' methods are another promising paradigm for collecting demonstrations with robot hands. These methods make human hands move in a manner close to how robot hands move. Some works adapt exoskeletons for stroke rehabilitation \cite{dai2023gripper,lu2023visual}. Another recent work introduces a wearable hand \cite{wei2024hirohand} for data collection, but does not equip the hand with force sensing. These works only demonstrate grasping motions, as opposed to the kinds of dexterous fingertip manipulation tasks we consider.

\noindent \rev{\textbf{Using forces to inform actions for imitation learning:}
DexForce's main contribution is using contact forces, measured during kinesthetic demonstrations, to directly compute robot actions for learning dexterous manipulation policies. A few prior works have also proposed methods that learn from kinesthetic demonstrations with force sensing on robot hands \cite{li2014learnimpedance,gokhan2023hapticexploration}, but these works assume that the robot always maintains non-slipping contact with the object. Consequently, they are only capable of basic, fixed-grasp object re-orientation motions. \cite{li2014learnimpedance} does use their method to demonstrate one task involving contact switching, but to achieve re-grasping, they need to use additional hand-designed heuristics tailored to one object. In contrast, DexForce applies to a much broader set of tasks because it does not assume a fixed grasp. In our experiments, we use \ourName{} for tasks that involve contact switching without additional heuristics.}

\rev{Using forces to inform actions for imitation learning has also been explored on robot arms. Some works use Dynamic Movement Primitives \cite{pastor2011onlinemovementadaptation, kormushev2011imitationlearning}, variable stiffness control \cite{kronader2014learningcompliant}, or hybrid force-position control \cite{liu2025forcemimic} to reproduce demonstration forces. Compared to DexForce, these methods have limited capabilities because they either do not allow for vision inputs \cite{pastor2011onlinemovementadaptation, kormushev2011imitationlearning, kronader2014learningcompliant} or use control primitives that only apply to some tasks \cite{liu2025forcemimic}.
Other, more recent, works use similar principles as \ourName{} to learn policies from kinesthetic demonstrations for robot arms \cite{hou2024adc, ablett2024multimodalforcematchedimitationlearning}. These works use their methods to open and close a door \cite{ablett2024multimodalforcematchedimitationlearning} or wipe a vase \cite{hou2024adc}. Unlike these works, which are restricted to tasks that can be done with robot arms, DexForce focuses on executing fine-grained tasks with robot hands. We further the findings of these works by showing that using measured forces from demonstrations to compute actions for policy learning is also valuable for fine-grained dexterous manipulation.}

\noindent \rev{\textbf{Forces as observations for imitation learning:}
We also study how including force data in observations affects policy performance. Many prior works have also studied the effects of including force and other tactile sensing modalities in policy observations, on both robot arms \cite{kim2024trainingrobotsrobotsdeep, edmonds2017feelingtheforce, aoyama2023fewshotlearningforcebasedmotions} and dexterous hands \cite{weinberg2024surveylearningbasedapproachesrobotic}. These works find that including force data in observations generally improves policy performance. Our findings, namely that using force data in observations never hurts policy performance but helps more for tasks requiring greater coordination and precision, contribute an additional nuanced insight to this body of work.}

\noindent \textbf{Open-loop strategies for in-hand manipulation:}
\ourName{} utilizes open-loop replay as part of its demonstration collection procedure. We are inspired by prior work that has shown the utility of open-loop strategies in complex dexterous manipulation \cite{wang2024penspin, bhatt2021surprisinglyrobust}.

\section{Method}\label{sec:method}

\begin{figure}[t]
    \centering
    \vspace{0.8mm}
    \includegraphics[width=0.9\linewidth]{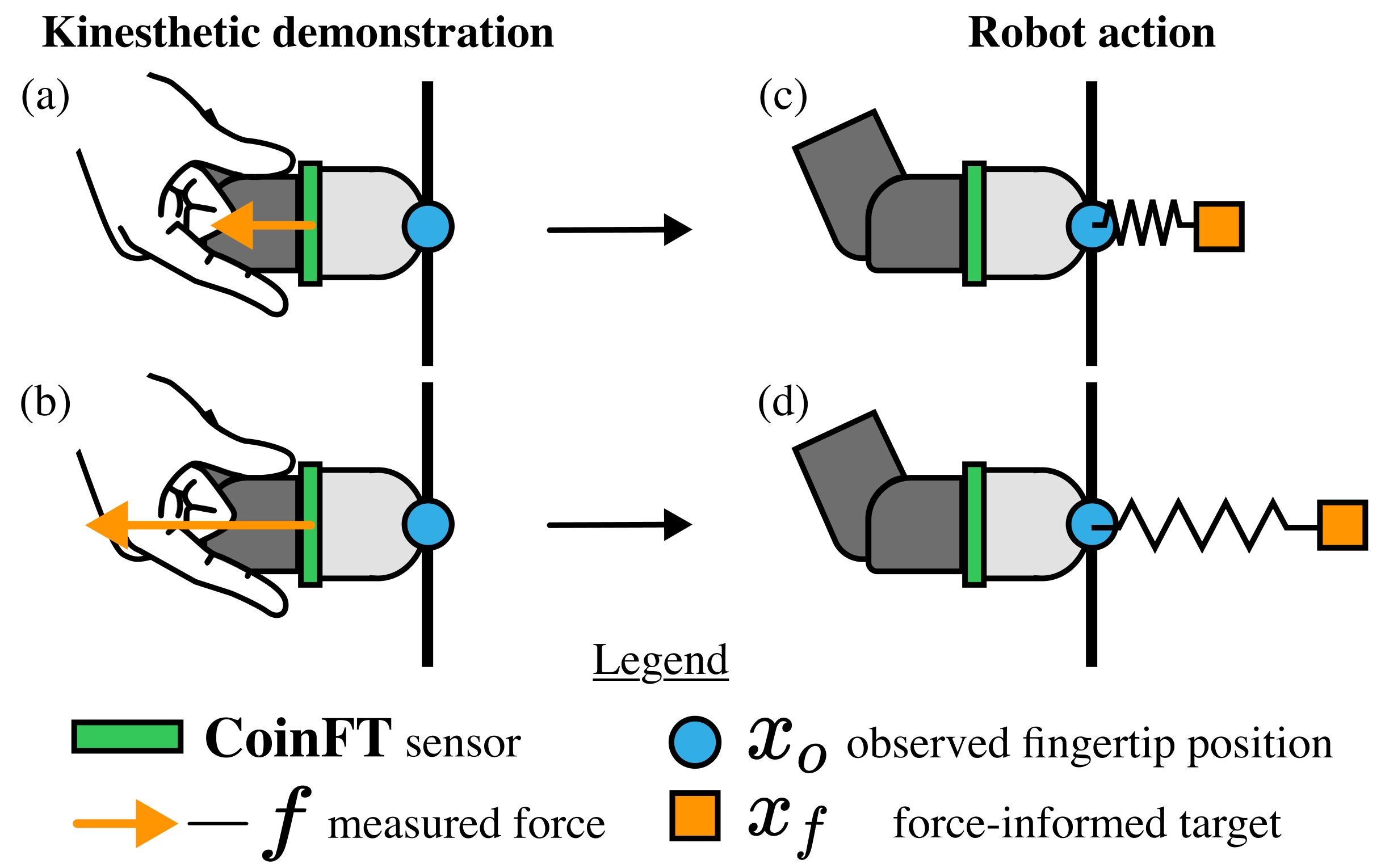}
    \caption{The left column shows kinesthetic demonstrations where the operator applies (a) a small force and (b) a larger force to a surface. In both scenarios, even though the operator applies different amounts of force, the current fingertip position recorded during the demonstrations, denoted with a blue circle, is the same. The difference between the observed fingertip positions and force-informed targets is proportional to the contact force $f$ (c, d). Note: we visualize $-f$, the contact force exerted by the object onto the finger.}
    \label{fig:method-spring}
    \vspace{-6mm}
\end{figure}

\ourName{} obtains demonstrations of contact-rich dexterous manipulation by extracting force-informed actions from kinesthetic demonstrations on a robot hand. To measure contact forces, we equip the robot hand with 6-axis force-torque sensors at the base of each fingertip. As illustrated in Fig. \ref{fig:teaser}a, with sensors at the base of each fingertip, the operator can firmly grip the robot finger anywhere above the sensor without impacting force measurements.

Kinesthetic demonstrations provide the robot state, consisting of observed fingertip positions $\mathbf{x}_{o}\in \mathbb{R}^3$ (blue circles in Fig. \ref{fig:method-spring}) and corresponding contact forces \mbox{$\mathbf{f}\in \mathbb{R}^3$} (orange arrows), for each finger. For tasks that require the robot to apply forces to an object, the observed fingertip positions cannot be directly used as actions because they would not reproduce measured forces. Intuitively, this is because, as illustrated in Figs. \ref{fig:method-spring}a and \ref{fig:method-spring}b, no matter how much force the operator applies with the robot finger via kinesthetic teaching, the observed fingertip position remains the same.

To obtain the right actions for policy learning, we use the observed fingertip positions and measured forces to compute force-informed position targets. When used as input to a Cartesian impedance controller, these force-informed position targets enable the robot to generate the right motions and forces to complete the task.

In the following sections, we will first describe the impedance controller. Then, we will present our method for extracting force-informed actions from kinesthetic demonstration data. Finally, we will present the complete two-stage \ourName{} demonstration collection procedure and how we use these demonstrations to learn policies.

\noindent \textbf{Cartesian impedance control:}
Impedance control \cite{hogan1984impedance} models the relationship between force and motion as a second-order mechanical system, characterized by a given mass, damping, and stiffness \cite{SicilianoKhatib2008springer}. It provides a means of producing desired motion and forces, instead of using explicit force control to directly track forces. Force control is notoriously unstable, especially when tasks involve discontinuous contact with the environment \cite{balachandran2017passivity}.

In a Cartesian impedance controller, the control force \mbox{$\mathbf{F} \in \mathbb{R}^3$} for one finger is given as 
\begin{equation}\label{eq:ctrl1}
    \mathbf{F} = \rev{k_p} (\mathbf{x}_d - \mathbf{x}_c) - \rev{k_v} ( \dot{\mathbf{x}}_c)
\end{equation}
where the first term scales the error between the desired and current fingertip positions $\mathbf{x}_d$ and $\mathbf{x}_c$, respectively, by a stiffness gain $k_p$, and the second term adds damping by scaling the fingertip velocity $\dot{\mathbf{x}}_c$ by the gain $k_v$. We tune the controller gains such that the fingertips follow a desired position trajectory while moving in free space (we use $k_p = 310$ and $k_v = 2$).
To apply this control force, we command the finger with joint torques $\tau$, given as
\begin{equation} \label{eq:ctrl2}
\tau = \mathbf{J}^T \mathbf{F} + \mathbf{g}
\end{equation}
where $\mathbf{J}$ is the finger Jacobian and $\mathbf{g}$ is a gravity compensation term.

\begin{figure*}[h]
    \centering
    \vspace{1mm}
    \includegraphics[width=0.85\linewidth]{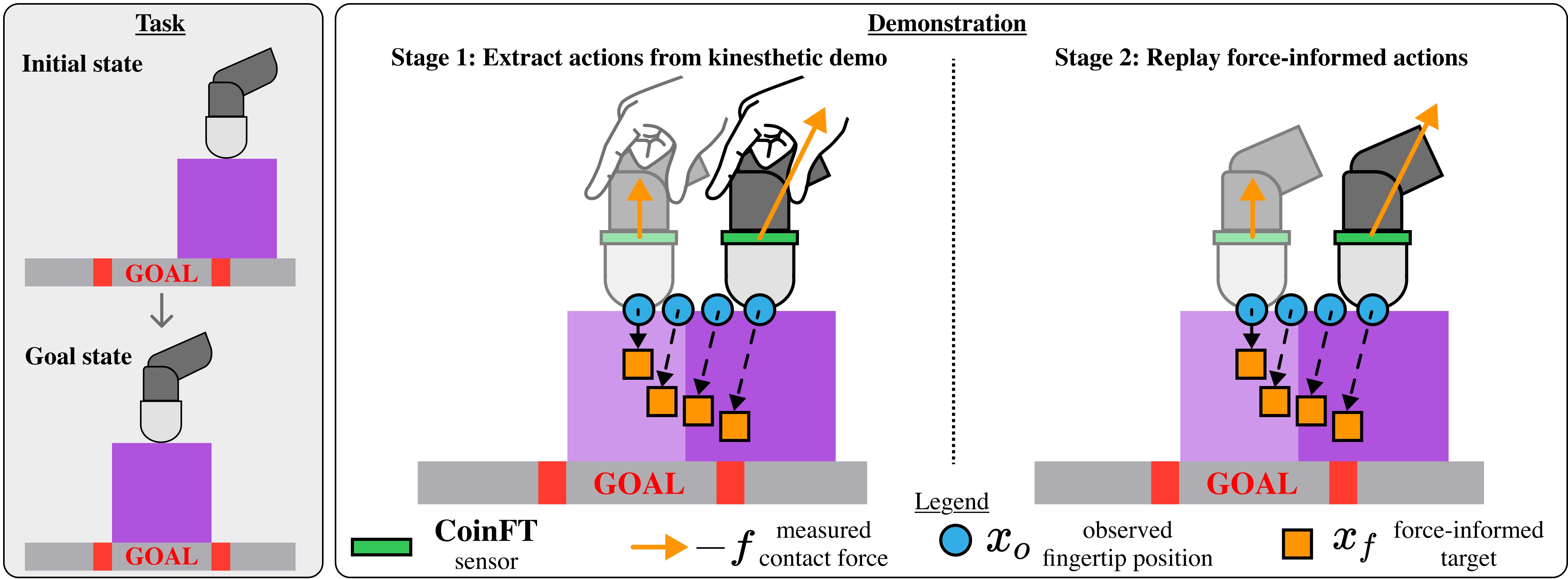}
    \caption{The \ourName{} two-stage demonstration collection procedure, illustrated with a one-finger task where the robot must slide a purple square along a fixed surface to the red goal region. Note: we visualize $-f$, the contact force exerted by the object onto the finger.}
    \label{fig:method}
    \vspace{-6mm}
\end{figure*}

\noindent \textbf{Computing force-informed position targets:}
Using impedance control enables a robot to apply forces by tracking desired positions. Given an observed fingertip position $\mathbf{x}_{o}$ and measured contact force $\mathbf{f}$ from a kinesthetic demonstration, our goal is to compute a desired fingertip position that produces $\mathbf{f}$ when tracked with the controller in  Eq. (\ref{eq:ctrl2}). We term this desired position a \textit{force-informed target}, $\mathbf{x}_f$.

The first term of the control law in Eq. (\ref{eq:ctrl1}) models the relationship between force and displacement as a spring. Assuming quasi-static motion, it follows that we can use the same model to compute a force-informed target $\mathbf{x}_f$ as a function of an observed fingertip position $\mathbf{x}_o$ and the contact force $\mathbf{f}$ exerted by the finger onto the object,
\begin{equation}\label{eq:force-to-desired}
    \mathbf{x}_f = \mathbf{x}_o + \rev{k_f} \mathbf{f}
\end{equation}
where $k_f$ is a hand-tuned scalar stiffness parameter. Intuitively, when an operator presses the robot fingertip onto a surface during a kinesthetic demonstration (Figs. \ref{fig:method-spring}a and \ref{fig:method-spring}b), the force-informed target will move farther into the surface as the applied force increases (Figs. \ref{fig:method-spring}c and \ref{fig:method-spring}d).
While Eq. (\ref{eq:force-to-desired}) does not model soft contact or friction, we empirically find it to be a sufficient model for many tasks.

\noindent \textbf{\ourName{} demonstration collection procedure} 
The \ourName{} demonstration collection procedure consists of two stages. Fig. \ref{fig:method} depicts this two-stage procedure on an example task where one finger must slide a square along a fixed rail into the red goal region.

In \textbf{Stage 1}, we extract force-informed targets from a kinesthetic demonstration.
From a kinesthetic demonstration, we record the robot state, consisting of a $T$ timestep-long trajectory of observed fingertip positions $\mathbf{x}_{o,1}...\mathbf{x}_{o,T}$ (blue circles in Fig. \ref{fig:method}) and corresponding contact forces \mbox{$\mathbf{f}_{1}...\mathbf{f}_{T}$} (orange arrows), for each finger. With this data, we compute a trajectory of force-informed position targets $\mathbf{x}_{f,1}...\mathbf{x}_{f,T}$ (orange squares) using Eq. (\ref{eq:force-to-desired}).

In \textbf{Stage 2}, we obtain the final demonstration used for policy learning. These demonstrations should also include images to give policies a visual understanding of the scene. However, images of a kinesthetic demonstration would contain the operator's hand, whereas test-time images only contain the robot. As such, the kinesthetic demonstrations collected in Stage 1 cannot be used directly to train policies. To obtain robot-only demonstrations that match test-time conditions, we replay the kinesthetic demonstration by tracking the trajectory of force-informed targets $\mathbf{x}_{f,1}...\mathbf{x}_{f,T}$, computed in Stage 1, with the Cartesian impedance controller in Eq. (\ref{eq:ctrl2}). We tune the stiffness $k_f$ in Eq. (\ref{eq:force-to-desired}) such that these replays successfully re-produce kinesthetic demonstrations. For all tasks, we use $k_f = 0.0045$.
Additionally, we use a wrist-mounted camera to capture image observations. From this replay, we record the new observed fingertip positions $\mathbf{x}^*_{o,1}...\mathbf{x}^*_{o,T}$, contact forces $\mathbf{f}^*_{1}...\mathbf{f}^*_{T}$, contact moments $\mathbf{m}^*_{1}...\mathbf{m}^*_{T}$, and \rev{wrist camera} images $\mathbf{I}_{1}...\mathbf{I}_{T}$.

\noindent \textbf{Policy learning:}
We use \ourName{} demonstrations to train Diffusion Policies \cite{chi2023diffusionpolicy}. An observation at timestep $t$ consists of image features of image $\mathbf{I}_t$ concatenated with the Stage 2 contact forces $\mathbf{f}^*_{t}$ and moments $\mathbf{m}^*_{t}$ of each finger. The policy outputs force-informed targets, which we execute with the impedance controller in Eq. (\ref{eq:ctrl2}). We supervise policy training on force-informed targets $\mathbf{x}_{f}$.

\vspace{-0.4mm}
\section{Experiments}

Our experiments answer the following three questions:

\noindent 1) How important is using \ourName{}'s force-informed actions to the success of imitation learning?

\noindent 2) How does including contact forces in policy observations impact policy performance?

\noindent \rev{3) How sensitive is DexForce to the stiffness parameter $k_f$?}

\begin{figure*}[h]
    \centering
    \vspace{2mm}
    \includegraphics[width=0.95\linewidth]{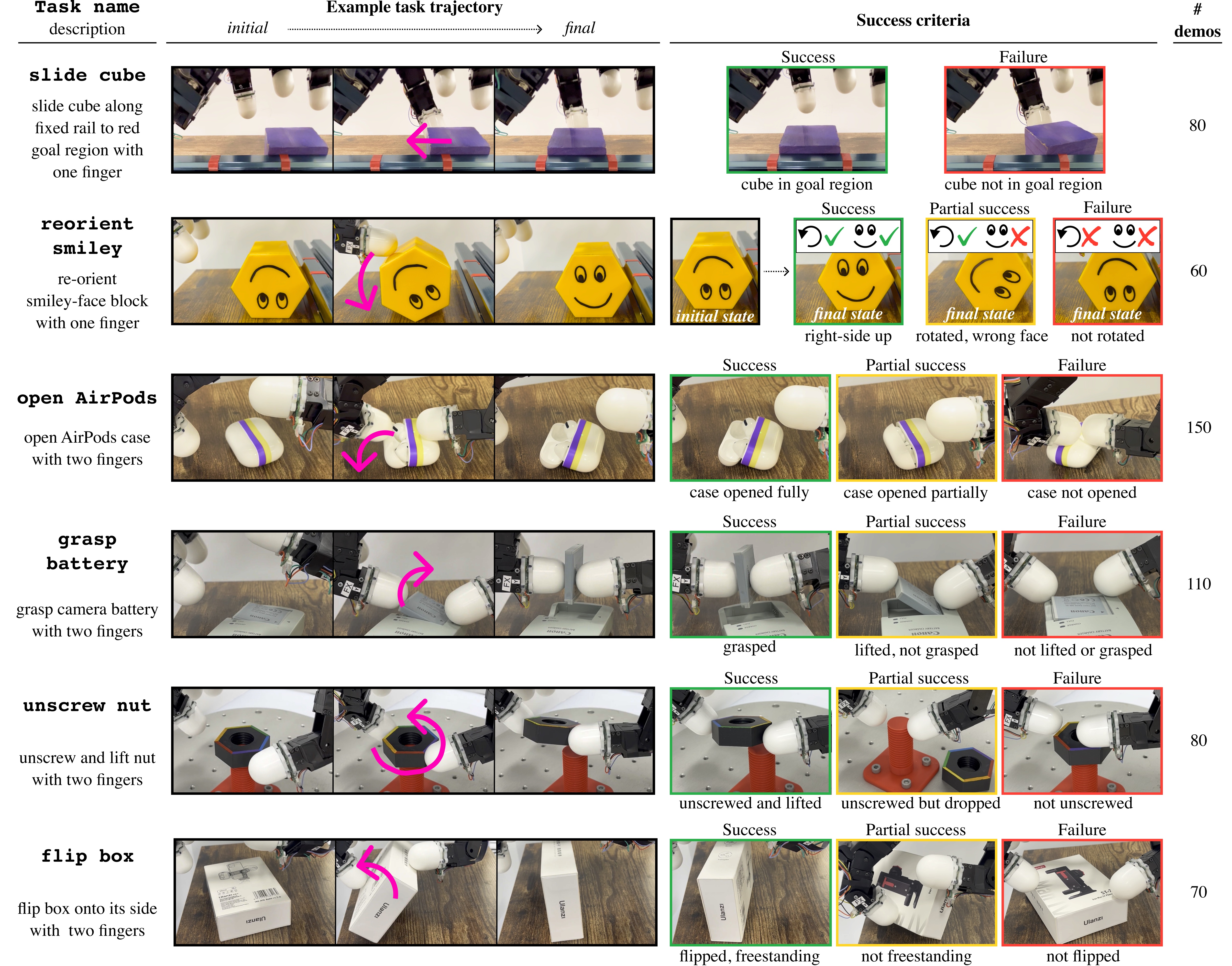}
    \caption{Task names, descriptions, example trajectories, success criteria, and number of training demonstrations.}
    \label{fig:tasks}
    \vspace{-7mm}
\end{figure*}

\begin{figure}[t]
    \centering
    \vspace{2mm}
    \includegraphics[width=0.8\linewidth]{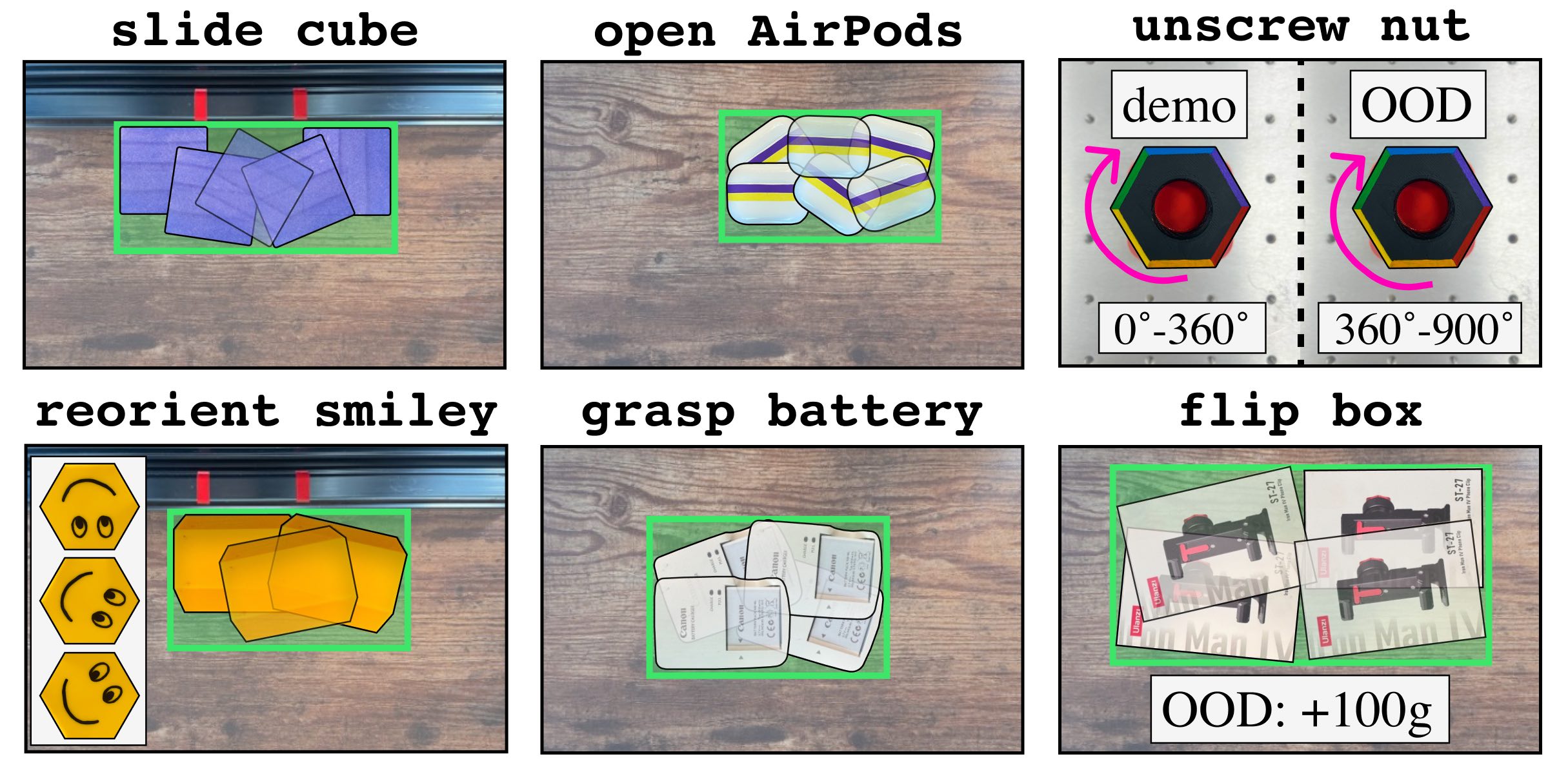}
    \caption{In our demonstrations, we randomly initialize object poses within the green regions. For our in-distribution evaluations, we sample 30 initial poses from within these green regions. For the \texttt{unscrew nut} demonstrations, we fix the position of the mount and randomize the initial rotation of the nut between 0\degree-360\degree. OOD scenarios: For \texttt{unscrew nut}, initial rotation of nut between 360\degree-900\degree. For \texttt{flip box}, the box is 100g heavier.
}
    \vspace{-8mm} %
    \label{fig:init-conf}
\end{figure}

\textbf{Tasks}: We evaluate policies on six contact rich manipulation tasks: \texttt{slide cube}, \texttt{reorient smiley}, \mbox{\texttt{open AirPods}}, \texttt{grasp battery}, \texttt{unscrew nut}, and \texttt{flip box}. Fig. \ref{fig:tasks} provides a description, an example trajectory, success criteria, and the number of training demonstrations for each task. We choose a variety of tasks that require different contact-rich interactions, like applying precise forces and making and breaking contact with the object. Because a human operator can comfortably collect kinesthetic demonstrations with up to two robot fingers, we consider \rev{one-finger extrinsic dexterity tasks} and two-finger tasks. In future work, we will explore mechanisms that enable an operator to collect kinesthetic demonstrations with more fingers.

In our evaluation, we label policy rollouts with one of three outcomes: success, failure, or partial success. Partial successes are scenarios where the robot makes progress towards completing the task but fails to apply the right forces to fully complete the task. For example, for \texttt{reorient smiley}, a partial success is when the robot has flipped the block at least once but failed to flip it to the correct orientation. We refer the reader to Fig. \ref{fig:tasks} for the partial success criteria of each task. Recording partial successes helps us understand failure modes of policies.

For each task, we vary the initial configuration of the object in both training demonstrations and evaluation scenes. As shown in Fig. \ref{fig:init-conf}, for \texttt{slide cube}, \texttt{reorient smiley}, \texttt{open AirPods}, \texttt{grasp battery}, and \texttt{flip box}, we vary the object pose within the reachable workspace of the Allegro hand. For \texttt{unscrew nut}, we vary how much the nut is screwed onto its mount.
To ensure consistency across all policy evaluations, within each task, we evaluate all policies on the same 30 random initial object configurations.

\begin{figure}[t]
    \centering
    \vspace{2mm}
    \includegraphics[width=0.9\linewidth]{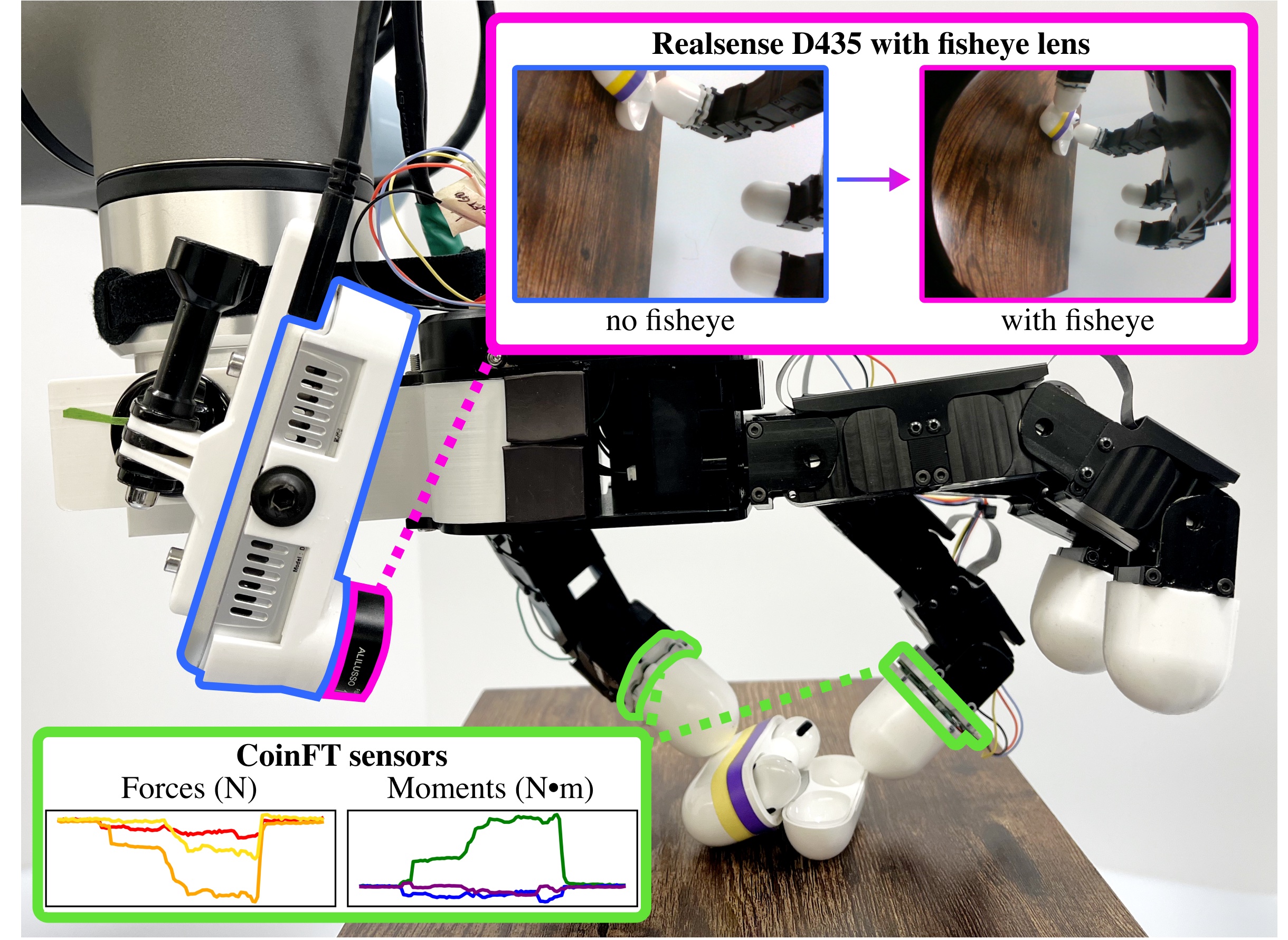}
    \caption{Hardware setup - Robot: Allegro hand; Camera: wrist-mounted RealSense D435 (blue), to capture RGB images, with a fisheye lens attachment (pink, adapted from \cite{wu2024tidybot}), which increases the camera field of view to capture the entire workspace of the hand; Force-torque sensors: Two CoinFT sensors \cite{choi2025coinftcoinsizedcapacitive6axis} (green) mounted at the bases of the thumb and index finger, to record 6-axis force-torque measurements. See project website for camera mounting and CoinFT details.}
    \label{fig:hw-setup}
    \vspace{-6mm}
\end{figure}

\textbf{Hardware setup:} We use an Allegro Hand instrumented with a wrist-mounted camera and 6-axis force-torque sensors \cite{choi2025coinftcoinsizedcapacitive6axis} at the bases of the thumb and index fingers, as shown in Fig. \ref{fig:hw-setup}. See the caption of Fig. \ref{fig:hw-setup} for more details.

\textbf{Policy learning details:} We train CNN-based Diffusion Policies \cite{chi2023diffusionpolicy} for all tasks, and use ResNet18 to encode RGB images. We use an observation horizon of 2, an action prediction horizon of 16, and an action execution horizon of 8. We train policies for 1000 epochs. We record demonstrations at 30Hz and downsample to 10Hz for training. For \texttt{slide cube}, \texttt{reorient smiley}, \texttt{open AirPods}, \texttt{unscrew nut}, and \texttt{flip box}, we use images of size [128, 128] pixels. To capture image details for \texttt{grasp battery}, we use [256, 256] pixels.

\noindent \textbf{Question 1) How important is using force-informed actions to the success of imitation learning?}
\begin{figure}[h]
    \centering
    \vspace{2mm}
    \includegraphics[width=0.98\linewidth]{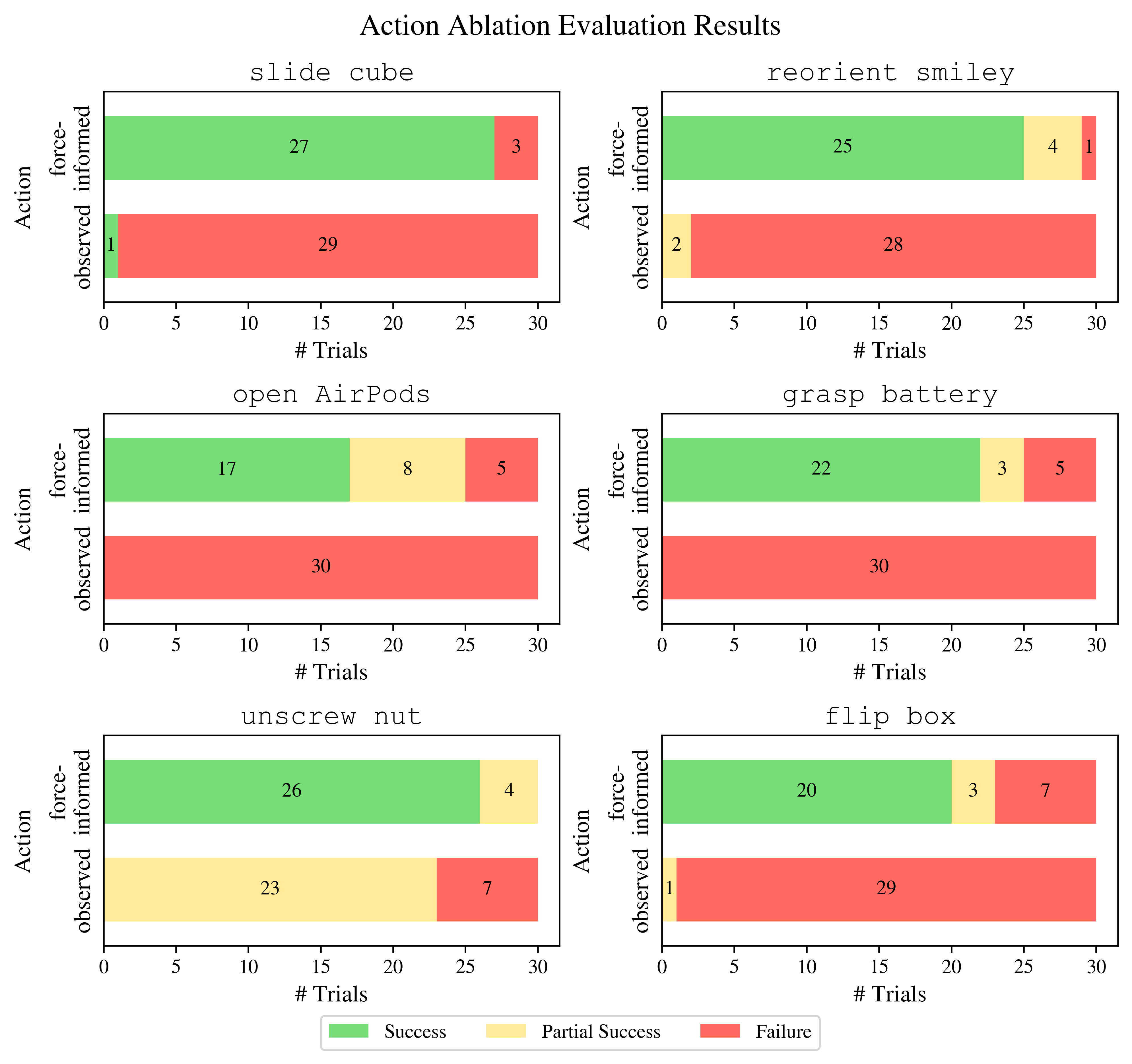}
    \caption{Using force-informed actions to train policies is critical for policy learning success. Policies trained directly on observed fingertip positions, which do not account for contact forces, have near-zero success rates.}
    \label{fig:plot-action}
    \vspace{-8mm} 
\end{figure}
For each task, we compare a policy trained on our force-informed targets to a policy trained directly on the observed fingertip positions, which do not account for contact forces. For policies trained on the observed fingertip positions, we use the next timestep's observed fingertip position as the action. All policies are trained with observations that consist of the RGB image features concatenated with the contact forces and moments for each finger.

As shown in Fig. \ref{fig:plot-action}, policies trained on our force-informed targets achieve success rates ranging from 90\% on \texttt{slide cube} to 57\% on \texttt{open AirPods}. Across all six tasks, the force-informed action policies achieve an average success rate of 76\%. In contrast, policies trained directly on observed fingertip positions have near-zero success rates. This stark difference in performance demonstrates the importance of computing actions based on measured contact forces.

Policies trained on observed fingertip positions fail because commanding the fingertips to go to those positions does not generate the right forces to complete tasks. For most tasks, the robot is able to move its fingers to the object but fails to apply any force, resulting in failures. On \texttt{unscrew nut}, the policy trained on observed fingertip positions achieves partial success on 23 out of 30 trials, where it is able to unscrew the nut. Yet, it never grasps the nut to fully complete the task. This suggests that force-informed actions are most important for the grasping part of the task. In policy rollouts, we observe that the hexagonal shape of the nut allows the robot to unscrew the nut by pushing on the corners instead of squeezing the nut to twist it. We refer the reader to our supplementary video for examples.

While the force-informed policy has the lowest success rate on \texttt{open AirPods}, achieving a full success on this task by opening the case completely is quite difficult. If the robot just partially opens the lid, the case will close itself under its own weight. For the case to remain open, the robot must apply the right forces to fully open the lid. We consider partially opening the case to be a partial success, because the robot still applies some meaningful force to the object. The force-informed policy achieves success on 17/30 trials and partial success on another 8 trials.

\noindent \textbf{Question 2: How does including contact forces in observations impact policy performance?}
In the prior section, we show that training on force-informed actions is critical for policy learning. Next, we ablate the inclusion of force data in policy observations to study how force observations impact policy performance. We supervise all policies on \ourName{}'s force-informed targets. For all tasks, we train three policies with the following observation ablations:

\noindent- \textbf{RGB, F/T}: RGB image features concatenated with the raw 6-axis contact force and moment vectors for each finger.

\noindent- \textbf{RGB, 0/1}: RGB image features concatenated with a binary contact signal for each finger. The contact signal is 1 if the finger is in contact (magnitude of the measured force exceeds 0.55N), and 0 if the finger is not in contact. We include this ablation because other work \cite{yin2023rotatingwithoutseeing} has shown that binary contact information is sufficient for dexterous manipulation.

\noindent- \textbf{RGB only}: RBG image features only. No force information.

\begin{figure}[h]
    \centering
    \vspace{2mm}
    \includegraphics[width=0.98\linewidth]{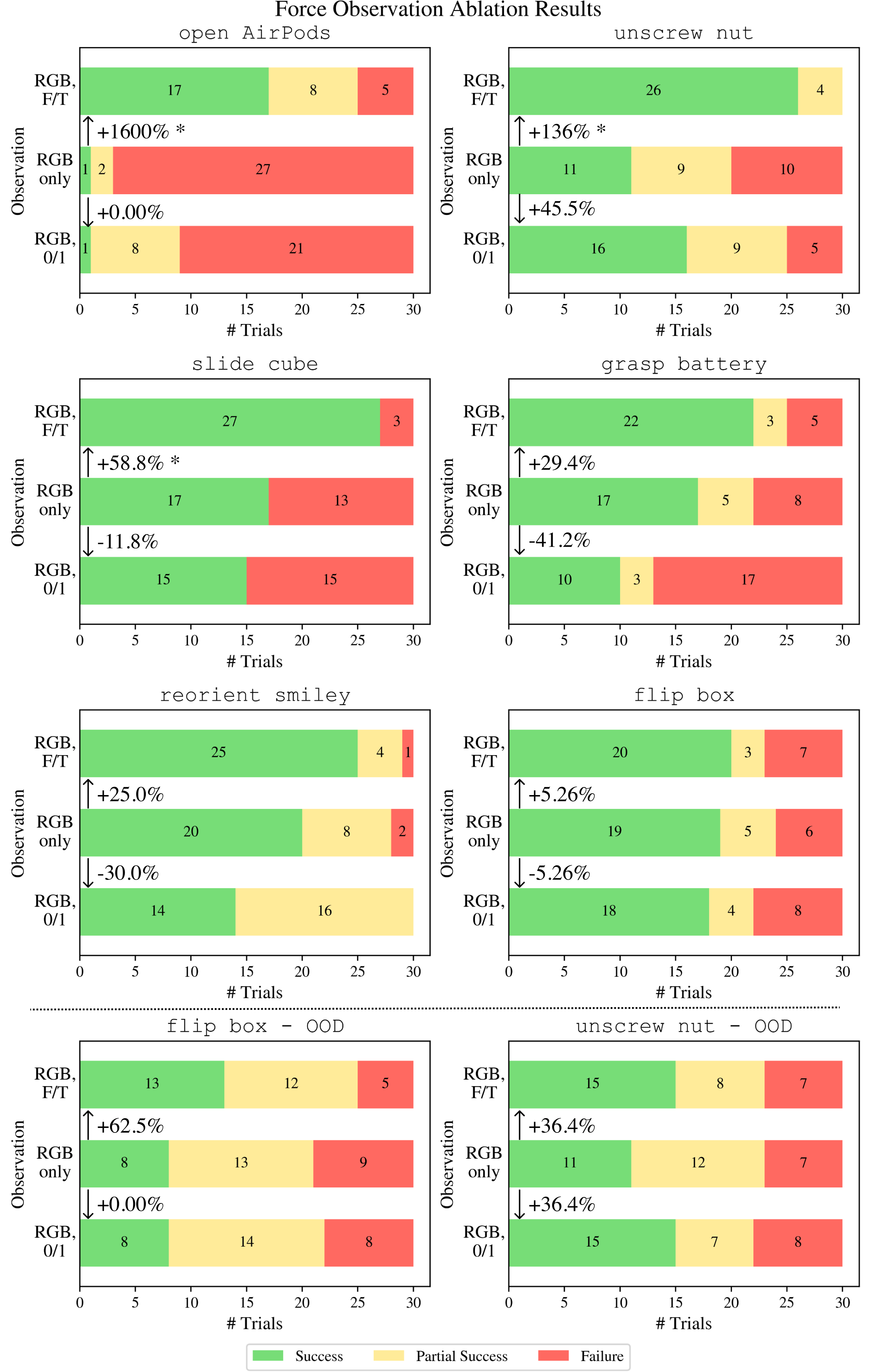}
    \caption{Force observation ablation results for six in-distribution evaluations (first three rows) and two out-of-distribution (OOD) evaluations (bottom row). In each plot, we show the success rate increase or decrease from \textbf{RGB-only} policies to \textbf{RGB, F/T} and \textbf{RGB, 0/1} policies (arrows between bars). \mbox{* denotes} statistical significance (Fisher's exact test, $\alpha$ = 0.05). For the in-distribution evaluations, plots are ordered from highest to lowest success rate increase from \textbf{RGB-only} to \textbf{RGB, F/T}. The \texttt{open AirPods}, \texttt{unscrew nut}, and \texttt{slide cube} tasks benefit the most from force observations.}
    \label{fig:plot-force}
    \vspace{-8mm} %
\end{figure}

Results are shown in Fig. \ref{fig:plot-force}. While including 6-axis force data in policy observations improves success rates for all tasks, it benefits \texttt{open AirPods}, \texttt{unscrew nut}, and \texttt{slide cube} the most. For these tasks, \textbf{RGB, F/T} policies have statistically significant success rate increases over \textbf{RGB-only} policies, improving performance on \texttt{open AirPods} by 1600\%, \texttt{unscrew nut} by 136\%, and \texttt{slide cube} by 58.8\%.

On these three tasks, \textbf{RGB-only} policies fail more often than \textbf{RGB, F/T} policies because the robot fails to apply forces that meet the required level of precision and coordination. For \texttt{open AirPods}, the \textbf{RGB-only} policy often pushes down on the lid of the case too hard and in the wrong direction, thus failing to move the lid at all. For \texttt{unscrew nut}, the \textbf{RGB-only} policy either fails to unscrew or grasp the nut. For \texttt{slide cube}, the \textbf{RGB-only} policy does not apply the right forces to slide the cube to the goal. We refer the reader to our supplementary video for examples.

We find that using binary contact information in policy observations is not comparable to using 6-axis contact force data; across all six tasks, \textbf{RGB, 0/1} policies have lower success rates than \textbf{RGB, F/T} policies. Furthermore, \textbf{RGB, 0/1} policies tend to do equivalently, or worse, than \textbf{RGB-only} policies, as shown in Fig. \ref{fig:plot-force}. This suggests that binary fingertip contact signals do not provide useful contact information to solve contact-rich tasks.

Finally, we test policies on out-of-distribution (OOD) variations of \texttt{flip box} and \texttt{unscrew nut}. We design OOD scenarios that are visually similar to demonstrations but have different force characteristics (see Fig. \ref{fig:init-conf} for more details). As expected, policies perform worse on OOD scenarios (last row of Fig. \ref{fig:plot-force}) than on their in-distribution counterparts. Even so, policies are able to successfully complete tasks about 50\% of the time, despite not being trained on these more challenging task variants. Furthermore, \textbf{RGB, F/T} policies continue to outperform \textbf{RGB-only} policies.

\noindent \rev{\textbf{Question 3: How sensitive is DexForce to the stiffness parameter $k_f$?}
For all six tasks, we conduct a sensitivity analysis of the stiffness parameter $k_f$ used to compute force-informed targets in Eq. (\ref{eq:force-to-desired}). Force-informed targets are first used in Stage 2 of our demonstration collection procedure, where we replay a kinesthetic demonstration by tracking a trajectory of force-informed targets. As such, a sufficient indicator of the feasibility of a given $k_f$ value is whether replaying a trajectory of force-informed targets, computed with that $k_f$, succeeds or fails to complete the task.}

\begin{figure}[h]
    \vspace{-1mm}
    \centering
    \includegraphics[width=0.98\linewidth]{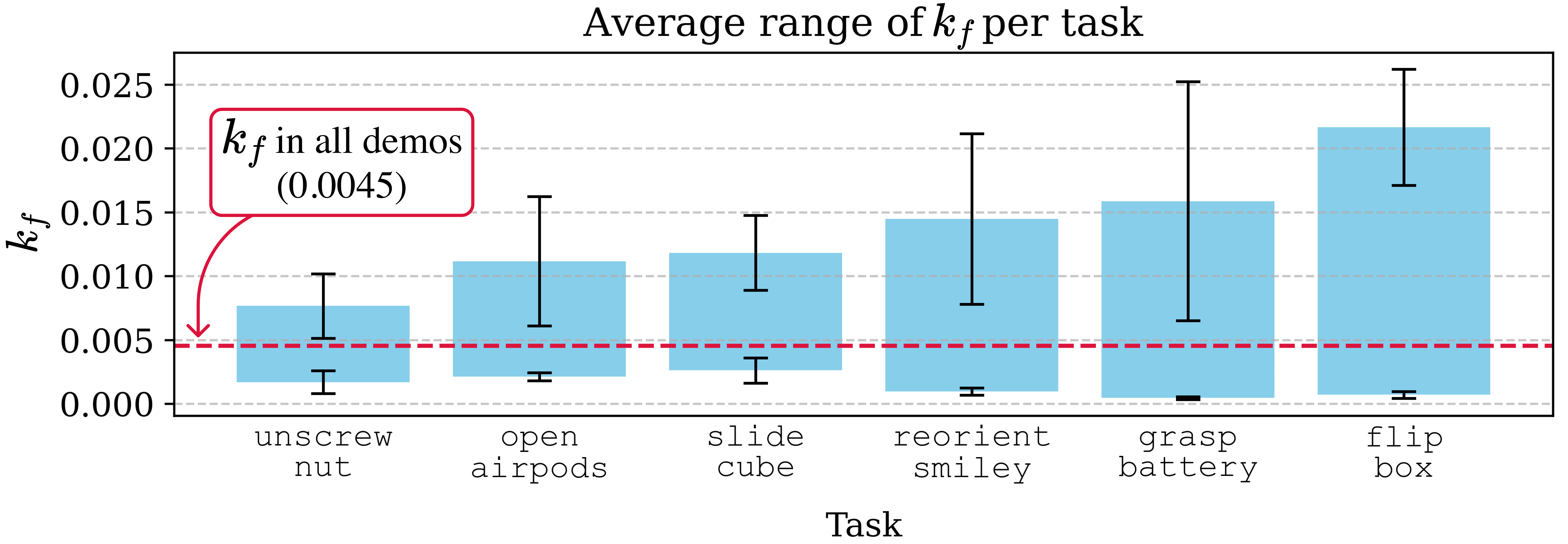}
    \caption{Sensitivity analysis of $k_f$ for each task. Each bar represents the average ranges of $k_f$ that result in successful \ourName{} demonstrations, across five unique kinesthetic demonstrations. We also show the standard deviations of the minimum and maximum $k_f$ values.}
    \label{fig:stiffness}
    \vspace{-3mm}
\end{figure}

\rev{For a given kinesthetic demonstration, we sweep across a range of $k_f$ values and conduct replay trials. For every trial, we select a value of $k_f$, compute the resulting force-informed targets with Eq. (\ref{eq:force-to-desired}), track this trajectory using our impedance controller, and label the trial as either a success or failure based on task success. We begin the sweep at $k_f = 0.0045$, which is the value we use to collect demonstrations for all our policy learning experiments. From here, we increase and decrease $k_f$ in 0.0001 increments until the replay fails. We set the upper limit of the sweep at $k_f = 0.025$ to avoid executing large, unsafe actions. For each task, we repeat this procedure for five different kinesthetic demonstrations.}

\rev{Fig. \ref{fig:stiffness} shows the average ranges of $k_f$ that result in successful replays. These results lead to three takeaways:}
 
\noindent \rev{1) $k_f$ does not need to be tuned for every task: 
Each task has a range of permissible $k_f$ values. This shows that \ourName{} is not overly sensitive to the choice of $k_f$. Furthermore, the ranges for all six tasks overlap with each other, showing that there is a range of $k_f$ values that work across different tasks. 
We also plot, as a red line, the value of $k_f = 0.0045$ that we use to collect all our demonstrations for policy learning, illustrating that this value falls comfortably within the permissible ranges of $k_f$ for each task.}

\noindent \rev{2) Contact-rich tasks can tolerate a range of force magnitudes:
For a given measured force from a kinesthetic demonstration, $\mathbf{f}$, the force applied by a force-informed target is proportional to $k_f$ (larger $k_f$ values correspond to actions with larger position displacements, which result in larger applied forces when executed with an impedance controller). Given this, each task's range of permissible $k_f$ values indicates that contact-rich tasks do not necessarily require applying exact forces. 
However, applied forces must still fall within the permissible range of magnitudes. Applying too little or too much force (i.e., a too small or too big $k_f$) will result in task failure. As an example, our supplementary video shows how applying too little or too much force results in failure of the \texttt{cube slide} task.
The crux of \ourName{} is that it leverages measured contact forces to compute actions that apply forces of both the right magnitudes and directions.
Finally, the minimum $k_f$ value is always greater than 0, reinforcing our main message that using measured forces to compute force-informed actions is critical.}

\noindent \rev{3) Some tasks are more force-sensitive than others:
While each of our six tasks has a range of permissible force magnitudes, some have smaller ranges than others, indicating that some tasks are more force-sensitive than others. Specifically, Fig. \ref{fig:stiffness} shows that \texttt{unscrew nut}, \texttt{open AirPods}, and \texttt{slide cube} have the smallest $k_f$ ranges. This indicates that applying the right forces is more important for these tasks than for the other tasks.
We think this is because \texttt{unscrew nut} involves many contact switches, \texttt{open AirPods} involves actuating an articulated object, and \texttt{slide cube} involves frictional interactions between the cube and the environment.
Additionally, in our force observation ablation results for each task, shown in Fig. \ref{fig:plot-force}, we see that the three most force-sensitive tasks are also the tasks that benefit most from having force data in policy observations. This suggests a correlation between the force-sensitivity of a task and the benefit of force observations.}

\section{Conclusion}

Using \ourName{}, we are able to collect robot demonstrations of contact-rich tasks that are difficult to demonstrate using current widely-used retargeting-based data collection methods. Policies trained on our force-informed actions achieve an average success rate of 76\% across six tasks, whereas policies trained on actions that do not account for contact forces have near-zero success rates.

While we acknowledge that \ourName{} may not be best suited for large-scale data collection, it shows how we can obtain actions that account for contact forces and that these force-informed actions are valuable for policy learning. We hope these insights not only enhance existing data collection methods \cite{chi2024universal, wei2024hirohand}, but also inform the design of future methods. Additionally, \ourName{} is a valuable research tool that enables us to study the impact of force-sensing on imitation learning for contact-rich dexterous manipulation. In future work, we will study alternative means of providing demonstrations that allow the operator to use full dexterity of robot and provide direct haptic feedback.

\bibliographystyle{IEEEtran}
\bibliography{references}

\end{document}